\documentclass[11pt,a4paper]{article}
\usepackage[hyperref]{acl2020}
\usepackage{times}
\usepackage{latexsym}

\usepackage{microtype}

\usepackage{amsfonts}
\usepackage{amsmath}
\usepackage{amssymb}
\usepackage{array}
\usepackage{url}
\usepackage{enumitem}
\usepackage{graphicx}
\usepackage{color}
\usepackage{transparent}
\usepackage{graphicx}
\usepackage{multirow}
\usepackage{siunitx}
\usepackage{subcaption}
\usepackage{pifont}
\usepackage{booktabs}
\usepackage[normalem]{ulem}
\usepackage{floatrow}

\aclfinalcopy 
\def\aclpaperid{2338} 

\newcolumntype{P}[1]{>{\centering\arraybackslash}p{#1}}
\newcolumntype{M}[1]{>{\centering\arraybackslash}m{#1}}
\newfloatcommand{capbtabbox}{table}[][\FBwidth]

\title{SKEP: Sentiment Knowledge Enhanced Pre-training \\ for Sentiment Analysis} 
\author{Hao Tian$^{\ddag,\dag}$, Can Gao$^{\dag}$, Xinyan Xiao$^{\dag}$, Hao Liu$^{\dag}$,  Bolei He$^{\dag}$,  \\
	\textbf{Hua Wu}$^{\dag}$\textbf{,} \textbf{Haifeng Wang}$^{\dag}$\textbf{,} \textbf{Feng Wu}$^{\ddag}$  \\
	$^{\dag}$Baidu Inc., Beijing, China \space\space\space\space  $^{\ddag}$University of Science and Technology of China\\
	\texttt{\{tianhao,gaocan01,xiaoxinyan,liuhao24,hebolei,wu\_hua,} \\
	\texttt{wanghaifeng\}@baidu.com, fengwu@ustc.edu.cn} 
}

\date{}

\begin{document}
\maketitle

\begin{abstract}
Recently, sentiment analysis has seen remarkable advance with the help of pre-training approaches. However, sentiment knowledge, such as sentiment words and aspect-sentiment pairs, is ignored in the process of pre-training, despite the fact that they are widely used in traditional sentiment analysis approaches. In this paper, we introduce Sentiment Knowledge Enhanced Pre-training (SKEP) in order to learn a unified sentiment representation for multiple sentiment analysis tasks. With the help of automatically-mined knowledge, SKEP conducts sentiment masking and constructs three sentiment knowledge prediction objectives, so as to embed sentiment information at the word, polarity and aspect level into pre-trained sentiment representation. In particular, the prediction of aspect-sentiment pairs is converted into multi-label classification, aiming to capture the dependency between words in a pair. Experiments on three kinds of sentiment tasks show that SKEP significantly outperforms strong pre-training baseline, and achieves new state-of-the-art results on most of the test datasets. We release our code at \url{https://github.com/baidu/Senta}. 
\end{abstract}

\section{Introduction}
Sentiment analysis refers to the identification of sentiment and opinion contained in the input texts that are often user-generated comments. In practice, sentiment analysis involves a wide range of specific tasks \cite{Bing2012Sentiment}, such as sentence-level sentiment classification, aspect-level sentiment classification, opinion extraction and so on. Traditional methods often study these tasks separately and design specific models for each task, based on manually-designed features \cite{Bing2012Sentiment} or deep learning \cite{zhang2018}.

Recently, pre-training methods \cite{peters2018deep,radford2018improving,devlin-etal-2019-bert,yang2019xlnet} have shown their powerfulness in learning general semantic representations, and have remarkably improved most natural language processing (NLP) tasks like sentiment analysis. These methods build unsupervised objectives at word-level, such as masking strategy \cite{devlin-etal-2019-bert}, next-word prediction \cite{radford2018improving} or permutation \cite{yang2019xlnet}. Such word-prediction-based objectives have shown great abilities to capture dependency between words and syntactic structures \cite{jawahar-etal-2019-bert}. However, as the sentiment information of a text is seldom explicitly studied, it is hard to expect such pre-trained general representations to deliver optimal results for sentiment analysis \cite{tang-etal-2014-learning}. 

Sentiment analysis differs from other NLP tasks in that it deals mainly with user reviews other than news texts. There are many specific sentiment tasks, and these tasks usually depend on different types of sentiment knowledge including sentiment words, word polarity and aspect-sentiment pairs. The importance of these knowledge has been verified by tasks at different level, for instance, sentence-level sentiment classification \cite{taboada2011lexicon,shin-etal-2017-lexicon,lei-etal-2018-multi}, aspect-level sentiment classification \cite{Vo_ijcai,zeng-etal-2019-variational},  opinion extraction \cite{li-lam-2017-deep,gui-etal-2017-question,fan-etal-2019-knowledge} and so on. Therefore, we assume that, by integrating these knowledge into the pre-training process, the learned representation would be more sentiment-specific and appropriate for sentiment analysis.

\begin{figure*}[t]
	\centering
	\scalebox{1.0}{
		\includegraphics[width=0.9\linewidth]{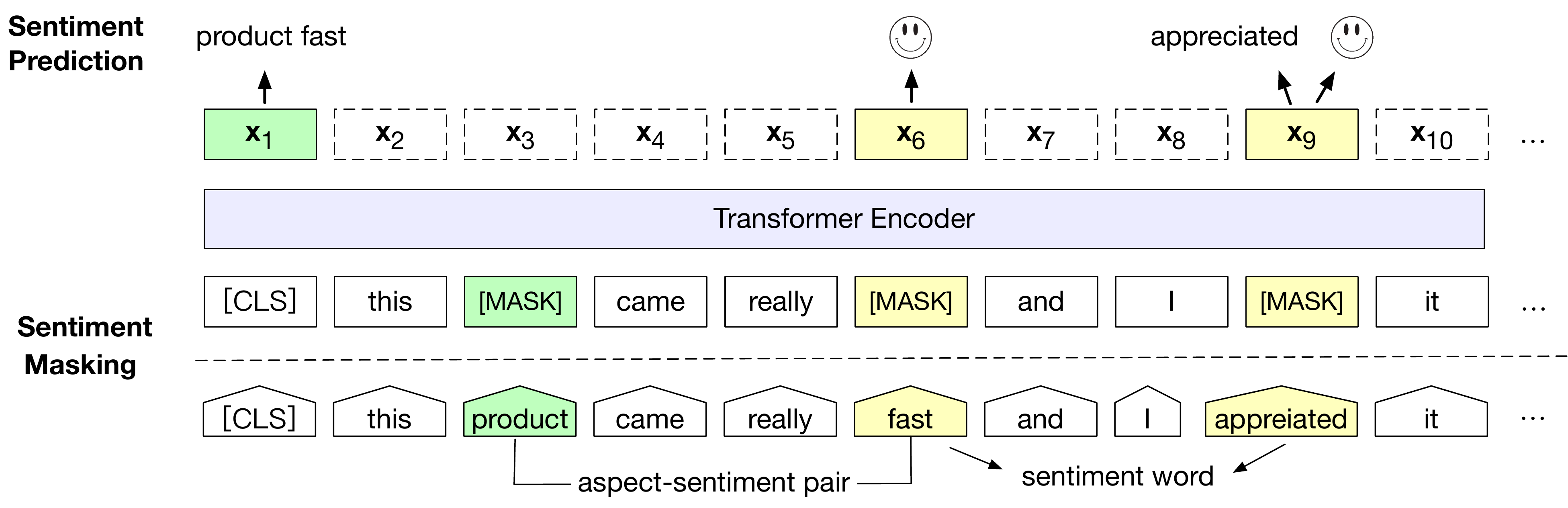}
	}
	\caption{Sentiment Knowledge Enhanced Pre-training (SKEP). SKEP contains two parts: (1) \textbf{Sentiment masking} recognizes the sentiment information of an input sequence based on automatically-mined sentiment knowledge, and produces a corrupted version by removing these informations. (2) \textbf{Sentiment pre-training objectives} require the transformer to recover the removed information from the corrupted version. The three prediction objectives on top are jointly optimized: Sentiment Word (SW) prediction (on $\mathbf{x}_9$), Word Polarity (SP) prediction (on $\mathbf{x}_6$ and $\mathbf{x}_9$), Aspect-Sentiment pairs (AP) prediction (on $\mathbf{x}_1$). Here, the smiley denotes positive polarity. Notably, on $\mathbf{x}_6$, only SP is calculated without SW, as its original word has been predicted in the pair prediction on $\mathbf{x}_1$.
 }
	\label{fig_skep} 
\end{figure*}

In order to learn a unified sentiment representation for multiple sentiment analysis tasks, we propose \emph{Sentiment Knowledge Enhanced Pre-training} (SKEP), where sentiment knowledge about words, polarity, and aspect-sentiment pairs are included to guide the process of pre-training. The sentiment knowledge is first automatically mined from unlabeled data (Section \ref{sec_mining}). With the knowledge mined, sentiment masking (Section \ref{sent_mask}) removes sentiment information from input texts. Then, the pre-training model is trained to recover the sentiment information with three sentiment objectives (Section \ref{sec_objetive}). 

SKEP integrates different types of sentiment knowledge together and provides a unified sentiment representation for various sentiment analysis tasks. This is quite different from traditional sentiment analysis approaches, where different types of sentiment knowledge are often studied separately for specific sentiment tasks. To the best of our knowledge, this is the first work that has tackled sentiment-specific representation during pre-training. Overall, our contributions are as follows:

\begin{itemize}
	\item We propose sentiment knowledge enhanced pre-training for sentiment analysis, which provides a unified sentiment representation for multiple sentiment analysis tasks. 
	\item Three sentiment knowledge prediction objectives are jointly optimized during pre-training so as to embed sentiment words, polarity, aspect-sentiment pairs into the representation. In particular, the pair prediction is converted into multi-label classification to capture the dependency between aspect and sentiment.
	\item SKEP significantly outperforms the strong pre-training methods RoBERTa \cite{liu2019roberta} on three typical sentiment tasks, and achieves new state-of-the-art results on most of the test datasets.
\end{itemize}

\section{Background: BERT and RoBERTa}
BERT \cite{devlin-etal-2019-bert} is a self-supervised representation learning approach for pre-training a deep transformer encoder \cite{Vaswani2017transformer}. BERT constructs a self-supervised objective called masked language modeling (MLM) to pre-train the transformer encoder, and relies only on large-size unlabeled data. With the help of pre-trained transformer, downstream tasks have been substantially improved by fine-tuning on task-specific labeled data. We follow the method of BERT to construct masking objectives for pre-training.

BERT learns a transformer encoder that can produce a contextual representation for each token of input sequences. In reality, the first token of an input sequence is a special classification token $\mathrm{[CLS]}$. In fine-tuning step, the final hidden state of $\mathrm{[CLS]}$ is often used as the overall semantic representation of the input sequence.

In order to train the transformer encoder, MLM is proposed. Similar to doing a cloze test, MLM predicts the masked token in a sequence from their placeholder. Specifically, parts of input tokens are randomly sampled and substituted. BERT uniformly selects $15\%$ of input tokens. Of these sampled tokens, $80\%$ are replaced with a special masked token $\mathrm{[MASK]}$, $10\%$ are replaced with a random token, $10\%$ are left unchanged. After the construction of this noisy version, the MLM aims to predict the original tokens in the masked positions using the corresponding final states.

Most recently, RoBERTa \cite{liu2019roberta} significantly outperforms BERT by robust optimization without the change of neural structure, and becomes one of the best pre-training models. RoBERTa also removes the next sentence prediction objective from standard BERT. To verify the effectiveness of our approach, this paper uses RoBERTa as a strong baseline.


\section{SKEP: Sentiment Knowledge Enhanced Pre-training}
We propose SKEP, Sentiment Knowledge Enhanced Pre-training, which incorporates sentiment knowledge by self-supervised training. As shown in Figure \ref{fig_skep}, SKEP contains sentiment masking and sentiment pre-training objectives. Sentiment masking (Section \ref{sent_mask})  recognizes the sentiment information of an input sequence based on automatically-mined sentiment knowledge (Section \ref{sec_mining}), and produces a corrupted version by removing this information. Three sentiment pre-training objectives (Section \ref{sec_objetive}) require the transformer to recover the sentiment information for the corrupted version. 

Formally, sentiment masking constructs a corrupted version $\widetilde{X}$ for an input sequence $X$ guided by sentiment knowledge $\mathcal{G}$.  $x_i$ and $\widetilde{x}_i$ denote the $i$-th token of $X$ and $\widetilde{X}$ respectively. After masking, a parallel data $(\widetilde{X}, X)$ is obtained. Thus, the transformer encoder can be trained with sentiment pre-training objectives that are supervised by recovering sentiment information using the final states of encoder $\widetilde{\mathbf{x}}_1, ... , \widetilde{\mathbf{x}}_n$.


\subsection{Unsupervised Sentiment Knowledge Mining}
\label{sec_mining} 
SKEP mines the sentiment knowledge from unlabeled data. As sentiment knowledge has been the central subject of extensive research, SKEP finds a way to integrate former technique of knowledge mining with pre-training. This paper uses a simple and effective mining method based on Pointwise Mutual Information (PMI) \cite{turney2002thumbs}. 

PMI method depends only on a small number of sentiment seed words and the word polarity $\mathrm{WP}(s)$ of each seed word $s$ is given. It first builds a collection of candidate word-pairs where each word-pair contains a seed word, 
and meet with pre-defined part-of-speech patterns as Turney \shortcite{turney2002thumbs}. Then, the co-occurrence of a word-pair is calculated by PMI as follows: 
\begin{align}
\mathrm{PMI}(w_1, w_2) = \log \dfrac{p(w_1,w_2)}{p(w_1) p(w_2)} 
\end{align}
Here, $p(.)$ denotes probability estimated by count. Finally, the polarity of a word is determined by the difference between its PMI scores with all positive seeds and that with all negative seeds.
\begin{align}
\mathrm{WP}(w) &= \sum_{\mathrm{WP}(s)=+} \mathrm{PMI}(w,s) \\ \nonumber
                        &-\sum_{\mathrm{WP}(s)=-} \mathrm{PMI}(w,s) 
\end{align}
If $\mathrm{WP}(w)$ of a candidate word $w$ is larger than $0$, then $w$ is a positive word, otherwise it is negative.

After mining sentiment words, aspect-sentiment pairs are extracted by simple constraints. An aspect-sentiment pair refers to the mention of an aspect and its corresponding sentiment word. Thus, a sentiment word with its nearest noun will be considered as an aspect-sentiment pair. The maximum distance between the aspect word and the sentiment word of a pair is empirically limited to no more than $3$ tokens.

Consequently, the mined sentiment knowledge $\mathcal{G}$ contains a collection of sentiment words with their polarity along with a set of aspect-sentiment pairs. Our research focuses for now the necessity of integrating sentiment knowledge in pre-training by virtue of a relatively common mining method. We believe that a more fine-grained method would further improve the quality of knowledge, and this is something we will be exploring in the nearest future.

\subsection{Sentiment Masking}
\label{sent_mask}
Sentiment masking aims to construct a corrupted version for each input sequence where sentiment information is masked. Our sentiment masking is directed by sentiment knowledge, which is quite different from previous random word masking. This process contains sentiment detection and hybrid sentiment masking that are as follows.


\paragraph{Sentiment Detection with Knowledge} Sentiment detection recognizes both sentiment words and aspect-sentiment pairs by matching input sequences with the mined sentiment knowledge $\mathcal{G}$.

\begin{enumerate}
	\item Sentiment Word Detection. The word detection is straightforward. If a word of an input sequence also occurs in the knowledge base $\mathcal{G}$, then this word is seen as a sentiment word.
	\item Aspect-Sentiment Pair Detection. The detection of an aspect-sentiment pair is similar to its mining described before. A detected sentiment word and its nearby noun word are considered as an aspect-sentiment pair candidate, and the maximum distance of these two words is limited to $3$. Thus, if such a candidate is also found in mined knowledge $\mathcal{G}$, then it is considered as an aspect-sentiment pair.  
\end{enumerate}

\paragraph{Hybrid Sentiment Masking} Sentiment detection results in three types of tokens for an input sequence: aspect-sentiment pairs, sentiment words and common tokens. The process of masking a sequence runs in following steps:
\begin{enumerate}
	\item Aspect-sentiment Pair Masking. At most $2$ aspect-sentiment pairs are randomly selected to mask. All tokens of a pair are replaced by $\mathrm{[MASK]}$ simultaneously. This masking provides a way for capturing the combination of an aspect word and a sentiment word.
	\item Sentiment Word Masking. For those un-masked sentiment words, some of them are randomly selected and all the tokens of them are substituted with $\mathrm{[MASK]}$ at the same time. The total number of tokens masked in this step is limited to be less than $10\%$.
	\item Common Token Masking. If the number of tokens in step 2 is insufficient, say less than $10\%$, this would be filled during this step with randomly-selected tokens. Here, random token masking is the same as RoBERTa.\footnote{For each sentence, we would always in total mask 10\% of its tokens at step 2 and 3. Among these masked tokens, 79.9\% are sentiment words (during step 2) and 20.1\% are common words (during step 3) in our experiment.}
\end{enumerate}

\subsection{Sentiment Pre-training Objectives}
\label{sec_objetive}
 Sentiment masking produces corrupted token sequences $\widetilde X$, where their sentiment information is substituted with masked tokens. Three sentiment objectives are defined to tell the transformer encoder to recover the replaced sentiment information. The three objectives, Sentiment Word (SW) prediction $L_{sw}$, Word Polarity (WP) prediction $L_{wp}$ and Aspect-sentiment Pair (AP) prediction $L_{ap}$ are jointly optimized. Thus, the overall pre-training objective $L$ is: 
\begin{align}
L = L_{sw} + L_{wp} + L_{ap}
\end{align}

\paragraph{Sentiment Word Prediction} Sentiment word prediction is to recover the masked tokens of sentiment words using the output vector $\widetilde{\mathbf x}_i$ from transformer encoder. $\widetilde{\mathbf x}_i$  is fed into an output softmax layer, which produces a normalized probability vector $\mathbf{\hat{y}}_i$ over the entire vocabulary. In this way, the sentiment word prediction objective $L_{sw}$ is to maximize the probability of original sentiment word $x_i$ as follows: 
\begin{align}
\mathbf{\hat{y}}_i &=  \mathrm{softmax} (\widetilde{\mathbf{x}}_i\mathbf{W}+\mathbf{b}) \\ 
L_{sw}  &= - \sum_{i=1}^{i=n} m_i \times  \mathbf{y}_i \log \mathbf{\hat y}_i 
\end{align}
Here, $\mathbf{W}$ and $\mathbf{b}$ are the parameters of the output layer. $m_i=1$ if $i$-th position of a sequence is masked sentiment word\footnote{In sentiment masking, we add common tokens to make up for the deficiency of masked tokens of sentiment words. $L_{sw}$ also calculates these common tokens, while $L_{wp}$ does not includes them.}, otherwise it equals to $0$. $\mathbf{y}_i$ is the one-hot representation of the original token $x_i$.

Regardless of a certain similarity to MLM of BERT, our sentiment word prediction has a different purpose. Instead of predicting randomly masking tokens, this sentiment objective selects those sentiment words for self-supervision. As sentiment words play a key role in sentiment analysis, the representation learned here is expected to be more suitable for sentiment analysis. 


\paragraph{Word Polarity Prediction} Word polarity is crucial for sentiment analysis. For example, traditional lexicon-based model \cite{turney2002thumbs} directly utilizes word polarity to classify the sentiment of texts. To incorporate this knowledge into the encoder, an objective called word polarity prediction $L_{wp}$ is further introduced. $L_{wp}$ is similar to $L_{sw}$. For each masked sentiment token ${\widetilde x}_i$, $L_{wp}$ calculated its polarity (positive or negative) using final state $\mathbf{\widetilde x}_i$.  Then the polarity of target corresponds to the polarity of the original sentiment word, which can be found from the mined knowledge.

\paragraph{Aspect-sentiment Pair Prediction} Aspect sentiment pairs reveal more information than sentiment words do. Therefore, in order to capture the dependency between aspect and sentiment, an aspect-sentiment pair objective is proposed. Especially, words in a pair are \emph{not} mutually exclusive. This is quite different from BERT, which assumes tokens can be independently predicted.


We thus conduct aspect-sentiment pair prediction with multi-label classification. We use the final state of classification token $\mathrm{[CLS]}$, which denotes representation of  the entire sequence, to predict pairs. $\mathrm{sigmoid}$ activation function is utilized, which allows multiple tokens to occur in the output at the same time. The aspect-sentiment pair objective $L_{ap}$ is denoted as follows:
\begin{align}
\mathbf{\hat{y}}_a &=  \mathrm{sigmoid} (\widetilde{\mathbf{x}}_1\mathbf{W}_{ap}+\mathbf{b}_{ap}) \\ 
L_{ap}  &=  - \sum_{a=1}^{a=A} \mathbf{y}_a \log \mathbf{\hat y}_a 
\end{align}
Here, $\mathbf{x}_1$ denotes the output vector of  $\mathrm{[CLS]}$. $A$ is the number of masked aspect-sentiment pairs in a corrupted sequence. $\mathbf{\hat{y}}_a$ is the word probability normalized by $\mathrm{sigmoid}$. $\mathbf{y}_a$ is the sparse representation of a target aspect-sentiment pair. Each element of $\mathbf{y}_a$ corresponds to one token of the vocabulary, and equals to $1$ if the target aspect-sentiment pair contains the corresponding token.\footnote{This means that the dimension of $\mathbf{y}_a$ equals to the vocabulary size of pre-training method, which is $50265$ in our experiment.} As there are multiple elements of $\mathbf{y}_a$ equals to $1$, the predication here is multi-label classification.\footnote{It is possible to predict masked pairs with CRF-layer. However, it is more than 10-times slower than multi-label classification, thus could not be used in pre-training. }

\section{Fine-tuning for Sentiment Analysis}
We verify the effectiveness of SKEP on three typical sentiment analysis tasks: sentence-level sentiment classification, aspect-level sentiment classification, and opinion role labeling. On top of the pre-trained transformer encoder, an output layer is added to perform task-specific prediction. The neural network is then fine-tuned on task-specific labeled data.

\paragraph{Sentence-level Sentiment Classification}  This task is to classify the sentiment polarity of an input sentence. The final state vector of classification token $\mathrm{[CLS]}$ is used as the overall representation of an input sentence. On top of the transformer encoder, a classification layer is added to calculate the sentiment probability based on the overall representation.

\paragraph{Aspect-level Sentiment Classification} This task aims to analyze fine-grained sentiment for an aspect when given a contextual text. Thus, there are two parts in the input: aspect description and contextual text. These two parts are combined with a separator $\mathrm{[SEP]}$, and fed into the transformer encoder. This task also utilizes the final state of the first token  $\mathrm{[CLS]}$ for classification.

\paragraph{Opinion Role Labeling} This task is to detect fine-grained opinion, such as holder and target, from input texts. Following SRL4ORL \cite{marasovic-frank-2018-srl4orl}, this task is converted into sequence labeling, which uses BIOS scheme for labeling, and a CRF-layer is added to predict the labels.\footnote{All the pretraining models, including our SKEP and baselines use CRF-Layer here, thus their performances are comparable.}

\begin{table}[t]
	\begin{center}
		\begin{tabular}{r c c c}
			\toprule
			Dataset         &Train    &Dev    &Test\\
			\hline
			SST-2            & 67k    & 872   & 1821 \\
			Amazon-2     & 3.2m   & 400k & 400k \\
			Sem-R     & 3608   & -       & 1120 \\ 
			Sem-L      & 2328   & -       & 638\\ 
			MPQA2.0        &287      & 100    & 95  \\
			\bottomrule
		\end{tabular}
	\end{center}
	\caption{Numbers of samples for each dataset. Sem-R and Sem-L refer to restaurant and laptop parts of SemEval 2014 Task 4. }
	\label{table_data}
\end{table}

\begin{table}[t]
	\begin{center}
		\begin{tabular}{r c c c}
			\toprule
			Dataset         &Learning Rate    &Batch    &Epoch\\
			\hline
			SST-2            & 1e-5, 2e-5, 3e-5   & {16, 32}   & 10 \\
			Amazon-2     & 2e-5, 5e-5   & 4 & 3 \\
			Sem-R           & 3e-5   & 16       & 5 \\ 
			Sem-L           & 3e-5   & 16       & 5\\ 
			MPQA2.0       &3e-5     & 16    & 5  \\
			\bottomrule
		\end{tabular}
	\end{center}
	\caption{Hyper-parameters for fine-tuning on each dataset. Batch and Epoch indicate batch size and maximum epoch respectively. }
	\label{table_finetune}
\end{table}

\begin{table*}[h]
	\begin{center}
		\begin{tabular}{l  l c c c c c}
			\toprule
			& \multicolumn{2}{c}{Sentence-Level}   & \multicolumn{2}{c}{Aspect-Level}  & \multicolumn{2}{c}{Opinion Role}  \\
			Model                           &SST-2   &Amazon-2  &Sem-L   &Sem-R   &MPQA-Holder  & MPQA-Target \\
			\hline
			Previous SOTA                     &\textbf{97.1}$_{*}^{1}$  &97.37$^{2}$  &81.35$^{3}$    &87.89$^{4}$  &83.67/77.12$^{5}$ &81.59/73.16$^{5}$  \\ 
			\hline
			RoBERTa$_{base}$               &94.9   &96.61  &78.11   &84.93  &81.89/77.34 &80.23/72.19    \\  
			RoBERTa$_{base}$ + SKEP  &96.7   &96.94  &81.32   &87.92   &84.25/79.03 &82.77/74.82\\
			\hline
			RoBERTa$_{large}$              &96.5   &97.33  &79.22   &85.88   &83.52/78.59 &81.74/75.87\\
			RoBERTa$_{large}$ + SKEP  &{97.0}  &\textbf{97.56}  &\textbf{81.47}   &\textbf{88.01} &\textbf{85.77}/\textbf{80.99} &\textbf{83.59}/\textbf{77.41} \\
			\bottomrule
		\end{tabular}
	\end{center}
	\caption{Comparison with RoBERTa and previous SOTA. For MPQA, here reports both binary-F1 and prop-F1 as \cite{marasovic-frank-2018-srl4orl}, which are split by a slash. The scores of previous SOTA come from:
		$^{1}$\cite{raffel2019exploring, Lan2019ALBERTAL}; $^{2}$\cite{uda}; $^{3}$\cite{sdgcn}; $^{4}$\cite{Rietzler2019AdaptOG}; $^{5}$\cite{marasovic-frank-2018-srl4orl}. The SOTA score of SST-2 is from GLUE leaderboard \cite{wang-etal-2018-glue} on December 1, 2019, and the system is based on ensemble-model.}
	\label{table_main}
\end{table*}

\begin{table*}[h]
	\begin{center}
		\begin{tabular}{l  c c c c c c}
			\toprule
			& \multicolumn{2}{c}{Sentence-Level}   & \multicolumn{2}{c}{Aspect-Level}  & \multicolumn{2}{c}{Opinion Role}  \\
			Model                           &SST-2 dev   &Amazon-2  &Sem-L   &Sem-R   &MPQA-Holder  & MPQA-Target \\
			\hline
			RoBERTa$_{base}$            &95.21   &96.61  &78.11  &84.93  & 81.89/77.34 & 80.23/72.19    \\ 
			\hline
			+ Random Token       &95.57   &96.73  &78.89  &85.77   &82.71/77.71 & 80.86/73.01    \\
			+ SW                               &96.38   &96.82  &80.13  &86.92  &82.95/77.63 &81.18/73.15   \\
			+ SW + WP                      &96.51   &96.87   &80.32  &87.25   &82.97/77.82 &81.09/73.24   \\
			+ SW + WP + AP             &96.87   &96.94   &81.32  &87.92   &84.25/79.03 &82.77/74.82   \\ 
			\hline
			+ SW + WP + AP-I       &96.89   &96.93   &81.19  &87.71   &84.01/78.36 & 82.69/74.36   \\ 
			\bottomrule
		\end{tabular}
	\end{center}
	\caption{Effectiveness of objectives. SW, WP, AP refers to pre-training objectives: Sentiment Word prediction, Word Polarity prediction and Aspect-sentiment Pair prediction. ``Random Token" denotes random token masking used in RoBERTa. AP-I denotes predicting words in an Aspect-sentiment Pair Independently.}
	\label{table_cmp}
\end{table*}

\section{Experiment}
\subsection{Dataset and Evaluation}
A variety of English sentiment analysis datasets are used in this paper. Table \ref{table_data} summarizes the statistics of the datasets used in the experiments. These datasets contain three types of tasks: (1) For sentence-level sentiment classification, Standford Sentiment Treebank (SST-2) \cite{socher-etal-2013-recursive} and Amazon-2 \cite{NIPS2015_5782} are used. In Amazon-2, $400k$ of the original training data are reserved for development. The performance is evaluated in terms of accuracy. (2) Aspect-level sentiment classification is evaluated on Semantic Eval 2014 Task4 \cite{pontiki-etal-2014-semeval}. This task contains both restaurant domain and laptop domain, whose accuracy is evaluated separately. (3) For opinion role labeling, MPQA 2.0 dataset \cite{wiebe2005annotating,pittir7563} is used. MPQA aims to extract the targets or the holders of the opinions. Here we follow the method of evaluation in SRL4ORL \cite{marasovic-frank-2018-srl4orl}, which is released and available online. 4-folder cross-validation is performed, and  the F-1 scores of both holder and target are reported. 

To perform sentiment pre-training of SKEP, the training part of Amazon-2 is used, which is the largest dataset among the list in Table \ref{table_data}. Notably, the pre-training only uses raw texts without any sentiment annotation. To reduce the dependency on manually-constructed knowledge and provide SKEP with the least supervision, we only use $46$ sentiment seed words. Please refers to the appendix for more details about seed words. 

\subsection{Experiment Setting}
We use RoBERTa \cite{liu2019roberta} as our baseline, which is one of the best pre-training models. Both base and large versions of RoBERTa are used. RoBERTa$_{base}$ and  RoBERTa$_{large}$ contain $12$ and $24$ transformer layers respectively. As the pre-training method is quite costly in term of GPU resources, most of the experiments are done on RoBERTa$_{base}$, and only the main results report the performance on RoBERTa$_{large}$. 

For SKEP, the transformer encoder is first initialized with RoBERTa, then is pre-trained on sentiment unlabeled data. An input sequence is truncated to 512 tokens. Learning rate is kept as $5e-5$, and batch-size is $8192$. The number of epochs is set to 3. For the fine-tuning of each dataset, we run 3 times with random seeds for each combination of parameters (Table \ref{table_finetune}), and choose the medium checkpoint for testing according to the performance on the development set.


\begin{table*}[t]{}
	\begin{center}{}
		\scalebox{0.8}{
			\begin{tabular}{l l  p{12cm}  c }
				\toprule
			From	& Model & Sentence Samples & Prediction \\
				
				\hline
			\multirow{4}{*}{SST-2} 	& \multirow{2}{*}{RoBERTa}  &  \setlength{\fboxsep}{0pt}
				{\colorbox[hsb]{0,0.4,1}{\textsl{altogether}}}
				{\colorbox[hsb]{0,0.4,1}{\textsl{,}}}
				{\colorbox[hsb]{0,0.4,1}{\textsl{this}}}
				{\colorbox[hsb]{0,0.4,1}{\textsl{is}}}
				{\colorbox[hsb]{0,0.2,1}{\textsl{\uwave{successful}}}}
				{\colorbox[hsb]{0,0.05,1}{\textsl{as}}}
				{\colorbox[hsb]{0,0.2,1}{\textsl{a}}}
				{\colorbox[hsb]{0,0.05,1}{\textsl{film}}}
				{\colorbox[hsb]{0,0.2,1}{\textsl{,}}}
				{\colorbox[hsb]{0,0.4,1}{\textsl{while}}}
				{\colorbox[hsb]{0,0.05,1}{\textsl{at}}}
				{\colorbox[hsb]{0,0.05,1}{\textsl{the}}}
				{\colorbox[hsb]{0,0.05,1}{\textsl{same}}}
				{\colorbox[hsb]{0,0.05,1}{\textsl{time}}}
				{\colorbox[hsb]{0,0.2,1}{\textsl{being}}}
				{\colorbox[hsb]{0,0.4,1}{\textsl{a}}}
				{\colorbox[hsb]{0,0.2,1}{\textsl{most}}}
				{\colorbox[hsb]{0,0.05,1}{\textsl{touching}}}
				{\colorbox[hsb]{0,0.05,1}{\textsl{reconsideration}}}
				{\colorbox[hsb]{0,0.05,1}{\textsl{of}}}
				{\colorbox[hsb]{0,0.05,1}{\textsl{the}}}
				{\colorbox[hsb]{0,0.05,1}{\textsl{familiar}}}
				{\colorbox[hsb]{0,0,1}{\textsl{\uwave{masterpiece}}}}
				{\colorbox[hsb]{0,0.2,1}{\textsl{.}}}
		& \multirow{2}{*}{positive}		\\

	\cline{2-4}	& \multirow{2}{*}{SKEP} &\setlength{\fboxsep}{0pt}
				{\colorbox[hsb]{0,0.2,1}{\textsl{altogether}}}
				{\colorbox[hsb]{0,0.2,1}{\textsl{,}}}
				{\colorbox[hsb]{0,0.2,1}{\textsl{this}}}
				{\colorbox[hsb]{0,0.2,1}{\textsl{is}}}
				{\colorbox[hsb]{0,0.4,1}{\textsl{\uwave{successful}}}}
				{\colorbox[hsb]{0,0.05,1}{\textsl{as}}}
				{\colorbox[hsb]{0,0.2,1}{\textsl{a}}}
				{\colorbox[hsb]{0,0.4,1}{\textsl{film}}}
				{\colorbox[hsb]{0,0.2,1}{\textsl{,}}}
				{\colorbox[hsb]{0,0.2,1}{\textsl{while}}}
				{\colorbox[hsb]{0,0.05,1}{\textsl{at}}}
				{\colorbox[hsb]{0,0.05,1}{\textsl{the}}}
				{\colorbox[hsb]{0,0.05,1}{\textsl{same}}}
				{\colorbox[hsb]{0,0.05,1}{\textsl{time}}}
				{\colorbox[hsb]{0,0.2,1}{\textsl{being}}}
				{\colorbox[hsb]{0,0.4,1}{\textsl{a}}}
				{\colorbox[hsb]{0,0.2,1}{\textsl{most}}}
				{\colorbox[hsb]{0,0.05,1}{\textsl{touching}}}
				{\colorbox[hsb]{0,0.05,1}{\textsl{reconsideration}}}
				{\colorbox[hsb]{0,0.05,1}{\textsl{of}}}
				{\colorbox[hsb]{0,0.05,1}{\textsl{the}}}
				{\colorbox[hsb]{0,0.05,1}{\textsl{familiar}}}
				{\colorbox[hsb]{0,0.5,1}{\textsl{\uwave{masterpiece}}}}
				{\colorbox[hsb]{0,0.2,1}{\textsl{.}}}
			& \multirow{2}{*}{positive}	
				\\
				\hline
				\hline
				
		\multirow{2}{*}{Sem-L} 	&	RoBERTa & \setlength{\fboxsep}{0pt}
				{\colorbox[hsb]{0,0.4,1}{\textsl{I}}}
				{\colorbox[hsb]{0,0.2,1}{\textsl{got}}}
				{\colorbox[hsb]{0,0.05,1}{\textsl{this}}}
				{\colorbox[hsb]{0,0.05,1}{\textsl{at}}}
				{\colorbox[hsb]{0,0.4,1}{\textsl{an}}}
				{\colorbox[hsb]{0,0.05,1}{\textsl{\uwave{amazing}}}}
				{\colorbox[hsb]{0,0.05,1}{\textsl{\uuline{price}}}}
				{\colorbox[hsb]{0,0,1}{\textsl{from}}}
				{\colorbox[hsb]{0,0.4,1}{\textsl{Amazon}}}
				{\colorbox[hsb]{0,0.05,1}{\textsl{and}}}
				{\colorbox[hsb]{0,0.4,1}{\textsl{it}}}
				{\colorbox[hsb]{0,0.05,1}{\textsl{arrived}}}
				{\colorbox[hsb]{0,0,1}{\textsl{just}}}
				{\colorbox[hsb]{0,0.05,1}{\textsl{in}}}
				{\colorbox[hsb]{0,0.05,1}{\textsl{time}}}
				{\colorbox[hsb]{0,0.4,1}{\textsl{.}}}
				& negative
				\\
				
		\cline{2-4}
			&	SKEP & \setlength{\fboxsep}{0pt}
				
				{\colorbox[hsb]{0,0.4,1}{\textsl{I}}}
				{\colorbox[hsb]{0,0.05,1}{\textsl{got}}}
				{\colorbox[hsb]{0,0.05,1}{\textsl{this}}}
				{\colorbox[hsb]{0,0.05,1}{\textsl{at}}}
				{\colorbox[hsb]{0,0.2,1}{\textsl{an}}}
				{\colorbox[hsb]{0,0.4,1}{\textsl{\uwave{amazing}}}}
				{\colorbox[hsb]{0,0.4,1}{\textsl{\uuline{price}}}}
				{\colorbox[hsb]{0,0,1}{\textsl{from}}}
				{\colorbox[hsb]{0,0.4,1}{\textsl{Amazon}}}
				{\colorbox[hsb]{0,0.05,1}{\textsl{and}}}
				{\colorbox[hsb]{0,0.4,1}{\textsl{it}}}
				{\colorbox[hsb]{0,0.05,1}{\textsl{arrived}}}
				{\colorbox[hsb]{0,0,1}{\textsl{just}}}
				{\colorbox[hsb]{0,0.05,1}{\textsl{in}}}
				{\colorbox[hsb]{0,0.05,1}{\textsl{time}}}
				{\colorbox[hsb]{0,0.2,1}{\textsl{.}}}
				& positive
				\\

				\bottomrule
			\end{tabular}
		}
	\end{center}
	\caption{Visualization of chosen samples. Words above wavy underline are mean sentiment words, and words above double underlines mean aspects. Color depth denotes importance for classification. The deeper color means more importance. The color depth is calculated by the attention weights with the classification token $\mathrm{[CLS]}$.}
	\label{table_vis}
\end{table*}

\begin{table}[t]
	\begin{center}
		\begin{tabular}{l c c c c }
			\toprule
			Model                             &SST-2 dev    &Sem-L  &Sem-R  \\
			\hline
			Sent-Vector          &96.87    &81.32  &87.92  \\ 
			Pair-Vector                &96.91    &81.38  &87.95  \\
			\bottomrule
		\end{tabular}
	\end{center}
	\caption{Comparison of vector used for aspect-sentiment pair prediction. Sent-Vector uses sentence representation (output vector of $\mathrm{[CLS]}$) for prediction, while pair-vector uses the concatenation of output vectors of the two words in a pair.
}
	\label{table_oc}
\end{table}

\subsection{Main Results}
We compare our SKEP method with the strong pre-training baseline RoBERTa and previous SOTA. The result is shown in Table \ref{table_main}.

Comparing with RoBERTa, SKEP significantly and consistently improves the performance on both base and large settings. Even on RoBERTa$_{large}$, SKEP achieves an improvement of up to $2.4$ points. According to the task types, SKEP achieves larger improvements on fine-grained tasks, aspect-level classification and opinion role labeling, which are supposed to be more difficult than sentence-level classification. We think this owes to the aspect-sentiment knowledge that is more effective for these tasks. Interestingly, ``RoBERTa$_{base}$ +  SKEP" always outperforms RoBERTa$_{large}$, except on Amazon-2. As the large version of  RoBERTa is computationally expensive, the base version of SKEP provides an efficient model for application. Compared with previous SOTA, SKEP achieves new state-of-the-art results on almost all datasets, with a less satisfactory result only on SST-2.

Overall, through comparisons of various sentiment tasks, the results strongly verify the necessity of incorporating sentiment knowledge for pre-training methods, and also the effectiveness of our proposed sentiment pre-training method.

\subsection{Detailed Analysis}
\paragraph{Effect of Sentiment Knowledge} SKEP uses an additional sentiment data for further pre-training and utilizes three objectives to incorporate three types of knowledge. Table \ref{table_cmp} compares the contributions of these factors. Further pre-training with random sub-word masking of Amazon, Roberta$_{base}$ obtains some improvements. This proves the value of large-size task-specific unlabeled data. However, the improvement is less evident compared with sentiment word masking. This indicates that the importance of sentiment word knowledge. Further improvements are obtained when word polarity and aspect-sentiment pair objectives are added, confirming the contribution of both types of knowledge. Compare ``+SW+WP+AP"  with ``+Random Token", the improvements are consistently significant in all evaluated data and is up to about $1.5$ points. 

Overall, from the comparison of objectives, we conclude that sentiment knowledge is helpful, and more diverse knowledge results in better performance. This also encourages us to use more types of knowledge and use better mining methods in the future.

\paragraph{Effect of Multi-label Optimization} Multi-label classification is proposed to deal with the dependency in an aspect-sentiment pair. To confirm the necessity of capturing the dependency of words in the aspect-sentiment pair, we also compare it with the method where the token is predicted independently, which is denoted by AP-I. AP-I uses softmax for normalization, and independently predicts each word of a pair as the sentiment word prediction. According to the last line that contains AP-I in Table \ref{table_cmp}, predicting words of a pair independently do not hurt the performance of sentence-level classification. This is reasonable as the sentence-level task mainly relies on sentiment words. In contrast, in aspect-level classification and opinion role labeling, multi-label classification is efficient and yields improvement of up to 0.6 points. This denotes that multi-label classification does capture better dependency between aspect and sentiment, and also the necessity of dealing with such dependency.

\paragraph{Comparison of Vector for Aspect-Sentiment Pair Prediction}  SKEP utilizes the sentence representation, which is the final state of classification token $\mathrm{[CLS]}$, for aspect-sentiment pair prediction. We call this Sent-Vector methods. Another way is to use the concatenation of the final vectors of the two words in a pair, which we call Pair-Vector. As shown in Table \ref{table_oc}, the performances of these two decisions are very close. We suppose this dues to the robustness of the pre-training approach. As using a single vector for prediction is more efficient, we use final state of token $\mathrm{[CLS]}$ in SKEP.

\paragraph{Attention Visualization} Table \ref{table_vis} shows the attention distribution of final layer for the [CLS] token when we adopt our SKEP model to classify the input sentences. On the SST-2 example, despite RoBERTa gives a correct prediction, its attention about sentiment is inaccurate. On the Sem-L case, RoBERTa fails to attend to the word ``amazing", and produces a wrong prediction. In contrast, SKEP produces correct predictions and appropriate attention of sentiment information in both cases. This indicates that SKEP has better interpretability.

\section{Related Work}

\paragraph{Sentiment Analysis with Knowledge}  Various types of sentiment knowledge, including sentiment words, word polarity, aspect-sentiment pairs, have been proved to be useful for a wide range of sentiment analysis tasks. 

Sentiment words with their polarity are widely used for sentiment analysis, including sentence-level sentiment classification \cite{taboada2011lexicon,shin-etal-2017-lexicon,lei-etal-2018-multi,barnes-etal-2019-lexicon}, aspect-level sentiment classification \cite{Vo_ijcai},  opinion extraction \cite{li-lam-2017-deep}, emotion analysis \cite{gui-etal-2017-question,fan-etal-2019-knowledge} and so on. Lexicon-based method \cite{turney2002thumbs, taboada2011lexicon} directly utilizes polarity of sentiment words for classification. Traditional feature-based approaches encode sentiment word information in manually-designed features to improve the supervised models \cite{pang2008opinion,agarwal-etal-2011-sentiment}. In contrast, deep learning approaches enhance the embedding representation with the help of sentiment words \cite{shin-etal-2017-lexicon}, or absorb the sentiment knowledge through linguistic regularization \cite{qian-etal-2017-linguistically,fan-etal-2019-knowledge}.

Aspect-sentiment pair knowledge is also useful for aspect-level classification and opinion extraction. Previous works often provide weak supervision by this type of knowledge, either for aspect-level classification \cite{zeng-etal-2019-variational} or for opinion extraction \cite{Yang2017TransferLF,Ding2017RecurrentNN}.

Although studies of exploiting sentiment knowledge have been made throughout the years, most of them tend to build a specific mechanism for each sentiment analysis task, so different knowledge is adopted to support different tasks. Whereas our method incorporates diverse knowledge in pre-training to provide a unified sentiment representation for sentiment analysis tasks.

\paragraph{Pre-training Approaches} Pre-training methods have remarkably improved natural language processing, using self-supervised training with large scale unlabeled data. This line of research is dramatically advanced very recently, and various types of methods are proposed, including ELMO \cite{peters2018deep}, GPT \cite{radford2018improving},  BERT \cite{devlin-etal-2019-bert}, XLNet \cite{yang2019xlnet} and so on. Among them, BERT pre-trains a bidirectional transformer by randomly masked word prediction, and have shown strong performance gains. RoBERTa \cite{liu2019roberta} further improves BERT by robust optimization, and become one of the best pre-training methods.



Inspired by BERT, some works propose fine-grained objectives beyond random word masking. SpanBERT \cite{joshi2019spanbert} masks the span of words at the same time. ERNIE \cite{sun2019} proposes to mask entity words. On the other hand, pre-training for specific tasks is also studied. GlossBERT \cite{huang-etal-2019-glossbert} exploits gloss knowledge to improve word sense disambiguation. SenseBERT \cite{levine2019sensebert} uses WordNet super-senses to improve word-in-context tasks. A different ERNIE \cite{Zhang2019ERNIE} exploits entity knowledge for entity-linking and relation classification.

 
\section{Conclusion}
In this paper, we propose Sentiment Knowledge Enhanced Pre-training for sentiment analysis. Sentiment masking and three sentiment pre-training objectives are designed to incorporate various types of knowledge for pre-training model. Thought conceptually simple, SKEP is empirically highly effective. SKEP significantly outperforms strong pre-training baseline RoBERTa, and achieves new state-of-the-art on most datasets of three typical specific sentiment analysis tasks. Our work verifies the necessity of utilizing sentiment knowledge for pre-training models, and provides a unified sentiment representation for a wide range of sentiment analysis tasks. 

In the future, we hope to apply SKEP on more sentiment analysis tasks, to further see the generalization of SKEP, and we are also interested in exploiting more types of sentiment knowledge and more fine-grained sentiment mining methods.


\section*{Acknowledgments}
We thanks Qinfei Li for her valuable comments. We also thank the anonymous reviewers for their insightful comments. This work was supported by the National Key Research and Development Project of China (No. 2018AAA0101900).

\bibliography{skep}

\begin{thebibliography}{42}
\expandafter\ifx\csname natexlab\endcsname\relax\def\natexlab#1{#1}\fi

\bibitem[{Agarwal et~al.(2011)Agarwal, Xie, Vovsha, Rambow, and
  Passonneau}]{agarwal-etal-2011-sentiment}
Apoorv Agarwal, Boyi Xie, Ilia Vovsha, Owen Rambow, and Rebecca Passonneau.
  2011.
\newblock \href {https://www.aclweb.org/anthology/W11-0705} {Sentiment analysis
  of twitter data}.
\newblock In \emph{Proceedings of the Workshop on Language in Social Media
  ({LSM} 2011)}.

\bibitem[{Barnes et~al.(2019)Barnes, Touileb, {\O}vrelid, and
  Velldal}]{barnes-etal-2019-lexicon}
Jeremy Barnes, Samia Touileb, Lilja {\O}vrelid, and Erik Velldal. 2019.
\newblock \href {https://www.aclweb.org/anthology/W19-6119} {Lexicon
  information in neural sentiment analysis: a multi-task learning approach}.
\newblock In \emph{Proceedings of the 22nd Nordic Conference on Computational
  Linguistics}.

\bibitem[{Devlin et~al.(2019)Devlin, Chang, Lee, and
  Toutanova}]{devlin-etal-2019-bert}
Jacob Devlin, Ming-Wei Chang, Kenton Lee, and Kristina Toutanova. 2019.
\newblock \href {https://www.aclweb.org/anthology/N19-1423} {{BERT}:
  Pre-training of deep bidirectional transformers for language understanding}.
\newblock In \emph{NAACL 2019}.

\bibitem[{Ding et~al.(2017)Ding, Yu, and Jiang}]{Ding2017RecurrentNN}
Ying Ding, Jianfei Yu, and Jing Jiang. 2017.
\newblock \href {https://aaai.org/ocs/index.php/AAAI/AAAI17/paper/view/14865}
  {Recurrent neural networks with auxiliary labels for cross-domain opinion
  target extraction}.
\newblock In \emph{AAAI 2017}.

\bibitem[{Fan et~al.(2019)Fan, Yan, Du, Gui, Bing, Yang, Xu, and
  Mao}]{fan-etal-2019-knowledge}
Chuang Fan, Hongyu Yan, Jiachen Du, Lin Gui, Lidong Bing, Min Yang, Ruifeng Xu,
  and Ruibin Mao. 2019.
\newblock \href {https://www.aclweb.org/anthology/D19-1563} {A knowledge
  regularized hierarchical approach for emotion cause analysis}.
\newblock In \emph{EMNLP 2019}.

\bibitem[{Gui et~al.(2017)Gui, Hu, He, Xu, Lu, and Du}]{gui-etal-2017-question}
Lin Gui, Jiannan Hu, Yulan He, Ruifeng Xu, Qin Lu, and Jiachen Du. 2017.
\newblock \href {https://www.aclweb.org/anthology/D17-1167} {A question
  answering approach for emotion cause extraction}.
\newblock In \emph{EMNLP 2017}.

\bibitem[{Huang et~al.(2019)Huang, Sun, Qiu, and
  Huang}]{huang-etal-2019-glossbert}
Luyao Huang, Chi Sun, Xipeng Qiu, and Xuanjing Huang. 2019.
\newblock \href {https://www.aclweb.org/anthology/D19-1355} {{G}loss{BERT}:
  {BERT} for word sense disambiguation with gloss knowledge}.
\newblock In \emph{EMNLP 2019}.

\bibitem[{Jawahar et~al.(2019)Jawahar, Sagot, and
  Seddah}]{jawahar-etal-2019-bert}
Ganesh Jawahar, Beno{\^\i}t Sagot, and Djam{\'e} Seddah. 2019.
\newblock \href {https://www.aclweb.org/anthology/P19-1356} {What does {BERT}
  learn about the structure of language?}
\newblock In \emph{ACL 2019}.

\bibitem[{Joshi et~al.(2019)Joshi, Chen, Liu, Weld, Zettlemoyer, and
  Levy}]{joshi2019spanbert}
Mandar Joshi, Danqi Chen, Yinhan Liu, Daniel~S. Weld, Luke Zettlemoyer, and
  Omer Levy. 2019.
\newblock {SpanBERT}: Improving pre-training by representing and predicting
  spans.
\newblock \emph{arXiv preprint arXiv:1907.10529}.

\bibitem[{Lan et~al.(2019)Lan, Chen, Goodman, Gimpel, Sharma, and
  Soricut}]{Lan2019ALBERTAL}
Zhen-Zhong Lan, Mingda Chen, Sebastian Goodman, Kevin Gimpel, Piyush Sharma,
  and Radu Soricut. 2019.
\newblock Albert: A lite bert for self-supervised learning of language
  representations.
\newblock \emph{ArXiv}, abs/1909.11942.

\bibitem[{Lei et~al.(2018)Lei, Yang, Yang, and Liu}]{lei-etal-2018-multi}
Zeyang Lei, Yujiu Yang, Min Yang, and Yi~Liu. 2018.
\newblock \href {https://www.aclweb.org/anthology/P18-2120} {A
  multi-sentiment-resource enhanced attention network for sentiment
  classification}.
\newblock In \emph{ACL 2018}.

\bibitem[{Levine et~al.(2019)Levine, Lenz, Dagan, Padnos, Sharir,
  Shalev-Shwartz, Shashua, and Shoham}]{levine2019sensebert}
Yoav Levine, Barak Lenz, Or~Dagan, Dan Padnos, Or~Sharir, Shai Shalev-Shwartz,
  Amnon Shashua, and Yoav Shoham. 2019.
\newblock \href {http://arxiv.org/abs/1908.05646} {Sensebert: Driving some
  sense into bert}.

\bibitem[{Li and Lam(2017)}]{li-lam-2017-deep}
Xin Li and Wai Lam. 2017.
\newblock \href {https://www.aclweb.org/anthology/D17-1310} {Deep multi-task
  learning for aspect term extraction with memory interaction}.
\newblock In \emph{EMNLP 2017}.

\bibitem[{Liu(2012)}]{Bing2012Sentiment}
Bing Liu. 2012.
\newblock Sentiment analysis and opinion mining.
\newblock In \emph{Synthesis Lectures on Human Language Technologies 5.1
  (2012): 1-167.}

\bibitem[{Liu et~al.(2019)Liu, Ott, Goyal, Du, Joshi, Chen, Levy, Lewis,
  Zettlemoyer, and Stoyanov}]{liu2019roberta}
Yinhan Liu, Myle Ott, Naman Goyal, Jingfei Du, Mandar Joshi, Danqi Chen, Omer
  Levy, Mike Lewis, Luke Zettlemoyer, and Veselin Stoyanov. 2019.
\newblock Roberta: A robustly optimized bert pretraining approach.
\newblock \emph{arXiv preprint arXiv:1907.11692}.

\bibitem[{Marasovi{\'c} and Frank(2018)}]{marasovic-frank-2018-srl4orl}
Ana Marasovi{\'c} and Anette Frank. 2018.
\newblock \href {https://www.aclweb.org/anthology/N18-1054} {{SRL}4{ORL}:
  Improving opinion role labeling using multi-task learning with semantic role
  labeling}.
\newblock In \emph{NAACL 2018}.

\bibitem[{Pang et~al.(2008)Pang, Lee et~al.}]{pang2008opinion}
Bo~Pang, Lillian Lee, et~al. 2008.
\newblock Opinion mining and sentiment analysis.
\newblock \emph{Foundations and Trends{\textregistered} in Information
  Retrieval}, 2(1--2):1--135.

\bibitem[{Peters et~al.(2018)Peters, Neumann, Iyyer, Gardner, Clark, Lee, and
  Zettlemoyer}]{peters2018deep}
Matthew~E Peters, Mark Neumann, Mohit Iyyer, Matt Gardner, Christopher Clark,
  Kenton Lee, and Luke Zettlemoyer. 2018.
\newblock Deep contextualized word representations.
\newblock \emph{arXiv preprint arXiv:1802.05365}.

\bibitem[{Pontiki et~al.(2014)Pontiki, Galanis, Pavlopoulos, Papageorgiou,
  Androutsopoulos, and Manandhar}]{pontiki-etal-2014-semeval}
Maria Pontiki, Dimitris Galanis, John Pavlopoulos, Harris Papageorgiou, Ion
  Androutsopoulos, and Suresh Manandhar. 2014.
\newblock \href {https://www.aclweb.org/anthology/S14-2004} {{S}em{E}val-2014
  task 4: Aspect based sentiment analysis}.
\newblock In \emph{{S}em{E}val 2014}.

\bibitem[{Qian et~al.(2017)Qian, Huang, Lei, and
  Zhu}]{qian-etal-2017-linguistically}
Qiao Qian, Minlie Huang, Jinhao Lei, and Xiaoyan Zhu. 2017.
\newblock \href {https://www.aclweb.org/anthology/P17-1154} {Linguistically
  regularized {LSTM} for sentiment classification}.
\newblock In \emph{ACL 2017}.

\bibitem[{Radford et~al.(2018)Radford, Narasimhan, Salimans, and
  Sutskever}]{radford2018improving}
Alec Radford, Karthik Narasimhan, Time Salimans, and Ilya Sutskever. 2018.
\newblock Improving language understanding with unsupervised learning.
\newblock Technical report, Technical report, OpenAI.

\bibitem[{Raffel et~al.(2019)Raffel, Shazeer, Roberts, Lee, Narang, Matena,
  Zhou, Li, and Liu}]{raffel2019exploring}
Colin Raffel, Noam Shazeer, Adam Roberts, Katherine Lee, Sharan Narang, Michael
  Matena, Yanqi Zhou, Wei Li, and Peter~J. Liu. 2019.
\newblock \href {http://arxiv.org/abs/1910.10683} {Exploring the limits of
  transfer learning with a unified text-to-text transformer}.

\bibitem[{Rietzler et~al.(2019)Rietzler, Stabinger, Opitz, and
  Engl}]{Rietzler2019AdaptOG}
Alexander Rietzler, Sebastian Stabinger, Paul Opitz, and Stefan Engl. 2019.
\newblock Adapt or get left behind: Domain adaptation through bert language
  model finetuning for aspect-target sentiment classification.
\newblock \emph{ArXiv}, abs/1908.11860.

\bibitem[{Shin et~al.(2017)Shin, Lee, and Choi}]{shin-etal-2017-lexicon}
Bonggun Shin, Timothy Lee, and Jinho~D. Choi. 2017.
\newblock \href {https://www.aclweb.org/anthology/W17-5220} {Lexicon integrated
  {CNN} models with attention for sentiment analysis}.
\newblock In \emph{Proceedings of the 8th Workshop on Computational Approaches
  to Subjectivity, Sentiment and Social Media Analysis}.

\bibitem[{Socher et~al.(2013)Socher, Perelygin, Wu, Chuang, Manning, Ng, and
  Potts}]{socher-etal-2013-recursive}
Richard Socher, Alex Perelygin, Jean Wu, Jason Chuang, Christopher~D. Manning,
  Andrew Ng, and Christopher Potts. 2013.
\newblock \href {https://www.aclweb.org/anthology/D13-1170} {Recursive deep
  models for semantic compositionality over a sentiment treebank}.
\newblock In \emph{EMNLP 2013}.

\bibitem[{Sun et~al.(2019)Sun, Wang, Li, Feng, Chen, Zhang, Tian, Zhu, Tian,
  and Wu}]{sun2019}
Yu~Sun, Shuohuan Wang, Yukun Li, Shikun Feng, Xuyi Chen, Han Zhang, Xin Tian,
  Danxiang Zhu, Hao Tian, and Hua Wu. 2019.
\newblock \href {https://arxiv.org/abs/1904.09223} {Ernie: Enhanced
  representation through knowledge integration}.
\newblock \emph{arXiv preprint arXiv:1904.09223}.

\bibitem[{Taboada et~al.(2011)Taboada, Brooke, Tofiloski, Voll, and
  Stede}]{taboada2011lexicon}
Maite Taboada, Julian Brooke, Milan Tofiloski, Kimberly Voll, and Manfred
  Stede. 2011.
\newblock Lexicon-based methods for sentiment analysis.
\newblock \emph{Computational linguistics}, 37(2):267--307.

\bibitem[{Tang et~al.(2014)Tang, Wei, Yang, Zhou, Liu, and
  Qin}]{tang-etal-2014-learning}
Duyu Tang, Furu Wei, Nan Yang, Ming Zhou, Ting Liu, and Bing Qin. 2014.
\newblock \href {https://www.aclweb.org/anthology/P14-1146} {Learning
  sentiment-specific word embedding for twitter sentiment classification}.
\newblock In \emph{ACL 2014}.

\bibitem[{Turney(2002)}]{turney2002thumbs}
Peter~D Turney. 2002.
\newblock Thumbs up or thumbs down?: semantic orientation applied to
  unsupervised classification of reviews.
\newblock In \emph{ACL 2002}.

\bibitem[{Vaswani et~al.(2017)Vaswani, Shazeer, Parmar, Uszkoreit, Jones,
  Gomez, Kaiser, and Polosukhin}]{Vaswani2017transformer}
Ashish Vaswani, Noam Shazeer, Niki Parmar, Jakob Uszkoreit, Llion Jones,
  Aidan~N. Gomez, Lukasz Kaiser, and Illia Polosukhin. 2017.
\newblock Attention is all you need.
\newblock In \emph{NIPS 2017}.

\bibitem[{Vo and Zhang(2015)}]{Vo_ijcai}
Duy-Tin Vo and Yue Zhang. 2015.
\newblock \href {http://dl.acm.org/citation.cfm?id=2832415.2832437}
  {Target-dependent twitter sentiment classification with rich automatic
  features}.
\newblock In \emph{IJCAI 2015}.

\bibitem[{Wang et~al.(2018)Wang, Singh, Michael, Hill, Levy, and
  Bowman}]{wang-etal-2018-glue}
Alex Wang, Amanpreet Singh, Julian Michael, Felix Hill, Omer Levy, and Samuel
  Bowman. 2018.
\newblock \href {https://doi.org/10.18653/v1/W18-5446} {{GLUE}: A multi-task
  benchmark and analysis platform for natural language understanding}.
\newblock In \emph{Proceedings of the 2018 {EMNLP} Workshop {B}lackbox{NLP}:
  Analyzing and Interpreting Neural Networks for {NLP}}.

\bibitem[{Wiebe et~al.(2005)Wiebe, Wilson, and Cardie}]{wiebe2005annotating}
Janyce Wiebe, Theresa Wilson, and Claire Cardie. 2005.
\newblock \href
  {https://www.microsoft.com/en-us/research/publication/annotating-expressions-of-opinions-and-emotions-in-language/}
  {Annotating expressions of opinions and emotions in language}.
\newblock \emph{Language Resources and Evaluation}.

\bibitem[{Wilson(2008)}]{pittir7563}
Theresa~Ann Wilson. 2008.
\newblock \href {http://d-scholarship.pitt.edu/7563/} {Fine-grained
  subjectivity and sentiment analysis: Recognizing the intensity, polarity, and
  attitudes of private states}.

\bibitem[{Xie et~al.(2019)Xie, Dai, Hovy, Luong, and Le}]{uda}
Qizhe Xie, Zihang Dai, Eduard~H. Hovy, Minh{-}Thang Luong, and Quoc~V. Le.
  2019.
\newblock \href {http://arxiv.org/abs/1904.12848} {Unsupervised data
  augmentation}.
\newblock \emph{CoRR}, abs/1904.12848.

\bibitem[{Yang et~al.(2019)Yang, Dai, Yang, Carbonell, Salakhutdinov, and
  Le}]{yang2019xlnet}
Zhilin Yang, Zihang Dai, Yiming Yang, Jaime Carbonell, Ruslan Salakhutdinov,
  and Quoc~V. Le. 2019.
\newblock \href {http://arxiv.org/abs/1906.08237} {Xlnet: Generalized
  autoregressive pretraining for language understanding}.

\bibitem[{Yang et~al.(2017)Yang, Salakhutdinov, and Cohen}]{Yang2017TransferLF}
Zhilin Yang, Ruslan Salakhutdinov, and William~W. Cohen. 2017.
\newblock Transfer learning for sequence tagging with hierarchical recurrent
  networks.
\newblock \emph{ArXiv}, abs/1703.06345.

\bibitem[{Zeng et~al.(2019)Zeng, Zhou, Liu, and
  Song}]{zeng-etal-2019-variational}
Ziqian Zeng, Wenxuan Zhou, Xin Liu, and Yangqiu Song. 2019.
\newblock \href {https://doi.org/10.18653/v1/N19-1036} {A variational approach
  to weakly supervised document-level multi-aspect sentiment classification}.
\newblock In \emph{NAACL 2019}.

\bibitem[{Zhang et~al.(2018)Zhang, Wang, and Liu}]{zhang2018}
Lei Zhang, Shuai Wang, and Bing Liu. 2018.
\newblock \href {https://doi.org/10.1002/widm.1253} {Deep learning for
  sentiment analysis : A survey}.
\newblock \emph{Wiley Interdisciplinary Reviews: Data Mining and Knowledge
  Discovery}.

\bibitem[{Zhang et~al.(2015)Zhang, Zhao, and LeCun}]{NIPS2015_5782}
Xiang Zhang, Junbo Zhao, and Yann LeCun. 2015.
\newblock \href
  {http://papers.nips.cc/paper/5782-character-level-convolutional-networks-for-text-classification.pdf}
  {Character-level convolutional networks for text classification}.
\newblock In \emph{NIPS 2015}.

\bibitem[{Zhang et~al.(2019)Zhang, Han, Liu, Jiang, Sun, and
  Liu}]{Zhang2019ERNIE}
Zhengyan Zhang, Xu~Han, Zhiyuan Liu, Xin Jiang, Maosong Sun, and Qun Liu. 2019.
\newblock \href {https://www.aclweb.org/anthology/P19-1139} {{ERNIE}: Enhanced
  language representation with informative entities}.
\newblock In \emph{ACL 2019}.

\bibitem[{Zhao et~al.(2019)Zhao, Hou, and Wu}]{sdgcn}
Pinlong Zhao, Linlin Hou, and Ou~Wu. 2019.
\newblock \href {http://arxiv.org/abs/1906.04501} {Modeling sentiment
  dependencies with graph convolutional networks for aspect-level sentiment
  classification}.
\newblock \emph{CoRR}, abs/1906.04501.

\end{thebibliography}
\bibliographystyle{acl_natbib}

\appendix 
\section{Appendix}
For sentiment knowledge mining, we construct $46$ sentiment seed words as follows. We first count the 9,750 items of \citet{qian-etal-2017-linguistically} on training data of Amazon-2, and get 50 most frequent sentiment words. Then, we manually filter out inappropriate words from these 50 words in a few minutes and finally get 46 sentiment words with polarities (Table \ref{table_senti_seeds}). The filtered words are \emph{need}, \emph{fun}, \emph{plot} and \emph{fine} respectively, which are all negative words.

\begin{table}[h]
	\begin{center}
		\begin{tabular}{| c | l |}
			\hline
			\multirow{5}{*}{\shortstack{positive\\word}}
			& great, good, like, just, will, well, \\
			& even, love, best, better, back, \\
			& want, recommend, worth, easy, \\
			& sound, right, excellent, nice, real, \\
			& fun, sure, pretty, interesting, stars  \\ 
			\hline
			\multirow{5}{*}{\shortstack{negative\\word}}               
			& too, little, bad, game, down,\\
			& long, hard, waste, disappointed, \\
			& problem, try,  poor, less, boring, \\
			& worst, trying, wrong, least,   \\
			& although, problems, cheap \\
			\hline
		\end{tabular}
	\end{center}
	\caption{Sentiment seed words used in our experiment. }
	\label{table_senti_seeds}
\end{table}

\end{document}